\documentclass[11pt,a4paper]{article}
\usepackage{times,latexsym}
\usepackage{url}
\usepackage[T1]{fontenc}
\usepackage{verbatim}
\usepackage{todonotes}
\usepackage{pifont}
\usepackage{booktabs}
\usepackage{stfloats}
\usepackage[hyperref]{acl2021}
\usepackage{microtype}
\usepackage{graphicx}

\usepackage{natbib}
\usepackage{todonotes}
\usepackage{url}
\usepackage{times}
\usepackage{float}

\aclfinalcopy % Uncomment this line for the final submission
\title{Which Argumentative Aspects of Hate Speech in Social Media\\can be reliably identified?}

\author{
  Damián Furman$^{1,2}$, 
  Pablo Torres$^{3}$,
  José A. Rodríguez$^{3}$,\\
  \textbf{Diego Letzen$^{3}$,
  Vanina Martínez$^{4}$,
  Laura Alonso Alemany$^{5,6}$}\\
  $^1$ Departamento de Computación, Universidad de Buenos Aires, Argentina\\
  $^2$ Consejo Nacional de Investigaciones Científicas y Técnicas, Argentina\\
  $^3$ Facultad de Filosofía, 
  Universidad Nacional de Córdoba, Argentina \\
  $^4$ Artificial Intelligence Research Institute (IIIA-CSIC), Barcelona, Spain\\
  $^5$ Facultad de Matemática, Astronomía y Física, Universidad Nacional de Córdoba, Argentina\\
  $^6$ Fundación Via Libre, Argentina
}

\date{}

\begin{document}
\maketitle
\begin{abstract}
With the increasing diversity of use cases of large language models, a more informative treatment of texts seems necessary. An argumentative analysis could foster a more reasoned usage of chatbots, text completion mechanisms or other applications. %, to complement the lack of generalization and reasoning that characterizes purely data-driven approaches. 
%, and to provide explanations for their analyses. %For applications like argument retrieval, stance detection, argument mining or counter-argumentation, this represents a crucial challenge.
%Many approaches have been proposed to analyze the argumentative properties of a text, 
However, it is unclear %how an analysis of argumentation can be successfully integrated in purely data-driven approaches. To begin with, it is unclear 
which aspects of argumentation can be reliably identified and integrated in language models.

In this paper we present an empirical assessment of the reliability with which different argumentative aspects can be automatically identified in hate speech in social media. We have enriched the Hateval corpus \cite{basile-etal-2019-semeval} with a manual annotation of some argumentative components, adapted from \citet{Wagemans2016ConstructingAP}'s Periodic Table of Arguments. %of text that is analytical, aims to represent important, theoretically fundamented, descriptively adequate and applicationwise useful aspects. 
We show that some components can be identified with reasonable reliability. For those that present a high error ratio, we analyze the patterns of disagreement between expert annotators and errors in automatic procedures, and we propose adaptations of those categories that can be more reliably reproduced.

%not included in the abstract: 
%We released an extension of Hateval corpus annotated with these components and an annotation manual, empirically assessed to which extent they can be manually and automatically identified in tweets, and discuss what this implies for the underlying theory and how this can be useful for further applications.
\end{abstract}

% %ORIGINAL ABSTRACT
% We present a framework for argumentative analysis of hate speech
% aimed to provide more useful insights into this phenomenon and facilitate intelligent downstream treatments.
% Based on Wagemann’s Periodic Table of Arguments, Besides justifications and conclusions, we also detect the core of the user’s reasoning, to be elaborated in subsequent automatic treatments, and identify domain-specific categories representing  characteristics of hate speech. We introduce the argumentative categories that we distinguish, as applied to Hateval corpus tweets, describe the results of manual annotation, and release both the annotated corpus and the annotation manual.
% We present several baselines for their automatic detection under different settings, including monolingual and multilingual showing that for most components, performance is close to human annotators.
% %SHORT, but that some components are easier to recognize than others.
% %SHORT We also evaluate performance under different sizes of training datasets and discuss the adecuacy of the published dataset for the task of automatic detection of each component.
% Finally, we analyze output examples and discuss ways for improving the recognition process.

\section{Introduction}

%With the rapid growth of social media platforms, hate speech and messages that were already present in our society found a way to spread faster and increase their reach.
%Laura: online bullying is correlated with offline bullying, but it is more widespread. Correlation with real life events 

With the impressive advances obtained in Large Language Models (LLMs), applications of automated language generation are quickly expanding to affect more and more areas of human activity, specially with the generalization of conversational chatbots. It is known that these models tend to amplify stereotypes, resulting in the naturalization of prejudices and finally the dehumanization of social groups in the form of hate speech.

Hate speech %in itself 
is a grave danger. The International Convention on the Elimination of all Forms of Racial Discrimination states that hate speech ``\textit{rejects basic human rights principles of human dignity and equality and seeks to degrade the position of individuals and groups in society's esteem}"\footnote{United Nations Strategy and Plan of Action on Hate Speech: Detailed Guidance on Implementation for United Nations Field Presences, 2020.}.
%Indeed, many legal systems typify some forms of hate speech as a crime. 

Through the amplification provided by social media and LLMs, its effects are also amplified, as it can deepen prejudice and stereotypes~\cite{citron2011intermediaries}. That is why great efforts have been made to detect and neutralize it. The most common form of neutralization to date has been banning hate speech from public forums. However, this strategy collisions with the right to freedom of expression. In addition, it is usually implemented by resorting to human moderators who are exposed to toxic content for long workdays.

Automatic argumentation analysis enables alternatives to censorship like argument retrieval and organization or automatic generation of counter-arguments. The recent developments of LLMs make these tasks more feasible. But, although they behave in a competent way from a purely conversational point of view, they do have not been designed to reason or argue. Moreover, they do not seem to be able to prevent harmful effects beyond very shallow guardrails, which is a critical concern when dealing with hate speech. That is why it seems necessary to enhance them beyond pure unannotated text, to obtain a more nuanced treatment of the argumentative dimension of texts. 

The question remains, \textit{how can we know which argumentative aspects will be useful for LLMs to improve their performance in nuanced, risky tasks like automatic generation of counter-arguments against hate speech in social media?}

% The predominant strategy adopted so far to counter hate speech in social media is to recognize, block and delete these messages and/or the users that generated it. This strategy has a disadvantage: blocking and deleting may prevent a hate message from spreading, but does not counter its consequences on those who were already reached by it. To do so, 
In this work we present the Argumentation Structure Of Hate Messages Online (ASOHMO), a protocol to annotate argumentative information in hate tweets, and an annotated dataset of tweets to train automatic classifiers. These annotations are an adaptation of~\citet{Wagemans2016ConstructingAP}'s proposal for hate speech in Twitter, where much of the argumentation refers to %. Our adaptation aims to achieve systematicity under a poorer argumentative context than more standard language, with many 
implicit elements, and one finds typos, incomplete phrases and incoherent syntax.
Despite this challenging context, by applying our protocol, we obtained substantial agreement between different human judges to identify the argumentative structure of tweets.
% Twitter requires an adaptation because it constitutes a poorer argumentative context than more standard language, with many implicit elements, full of incomplete phrases and incoherent syntax. Our proposal also aims to
% \todo{En esta parte aclararía que la dificultad en el proceso de anotación consiste en lograr un proceso que sea sencillo para que anotar no conlleve un esfuerzo extraordinario pero que a la vez tiene que permitir generar una cantidad de ejemplos suficientes que permitan aprender a un algoritmo el máximo posibe a partir de los ejemplos. Esto resulta ademas aún más dificil en el contexto de twitter. En este trabajo vamos a analizar además, cuál es la cantidad necesaria de ejemplos anotados para cada componente que permita alcanzar la máxima performance de un algoritmo}
% \xout{We believe that this tool could provide a better understanding of the arguments used to support xenophobous and racist conclusions, help humans to write better responses to those tweets and also help in the development of automatic counter-narrative generation tools.}
%Our proposal also aims for models that perform well without the need of  huge training datasets that require a prohibitive annotation effort. 
We also found that LLMs can successfully detect some of these 
argumentative components, even when few annotated examples are provided, which seems to indicate that it is feasible to finetune them to address some specific argumentation tasks and domains.
%progressivelly incrementing the size of the training datasets and analyzed variations in performance. We found that some of them can be satisfactorily identified achieving the best performance with a small dataset.

% and analyzed the impact of the size of the training datasets and assessed the relation between performance and annotation efforts.

% We also trained machine learning algorithms to detect argumentative components, and found that some of them can be satisfactorily identified, even when trained with a small dataset.

The rest of the paper is organized as follows. In the next Section we discuss relevant work, including the foundational \citet{Wagemans2016ConstructingAP}'s proposal. Section \ref{sec:framework}  describes the categories that we distinguish in our annotation framework, and in Section \ref{sec:corpus} we present how they apply to hate speech in social media, more concretely, to the manual annotation of the Hateval corpus \cite{basile-etal-2019-semeval}. Finally, in Section \ref{sec:automatic} we show how LLMs can identify some argumentative components, but not others, with varying degrees of success. We analyze the causes of low success and propose how to adapt the definition of the target argumentative aspects to improve their reliability of annotation, both manual and automatic. %analyze some experiments on the automatic identification of arguments in tweets and in Section \ref{sec:discusion} we discuss results and  future work. 
%\xout{Thus we believe it is feasible  We provide an analysis of difficulties in the annotation process and try to assess to which extent it is possible to systematize it. We also present experiments that show varying degrees of difficulty to automatically identify different argument components.}%SHORT, allowing to make decisions as to how they might be used for downstream applications.

%The rest of the paper is organized as follows. 
%In the following section, we discuss relevant work on argument annotation schemes and annotated corpora. Then we describe our annotation scheme and provide results for automatic identification of argumentation and insights gained from difficulties and discrepancies between annotators. Finally, we discuss possible applications of the proposed argumentative information scheme to other tasks, such as automatic counter-narratives generation.
\vspace{-0.05cm}
\section{Relevant Work}

There are many different proposals on how to model the argumentative aspects of texts, even if we only consider those aimed or used for computational application. We are not providing an exhaustive overview of approaches here, but just some examples to motivate and frame the model of argument that we present in this work.

One of the main distinctions between proposals is whether they are general purpose or domain specific. Domain-specific approaches propose tailored categories, like \citet{teufel-etal-1999-annotation}'s "\textit{background}", "\textit{aim}" or "\textit{comparison}" for scientific papers, or \citet{al-khatib-etal-2016-news}'s "\textit{anecdote}" or "\textit{statistics}" for the argumentative analysis of editorials. They tend to achieve good inter-annotator agreement and good accuracy in automatic identification, but are not portable to different domains.

General-purpose argumentation models have very different approaches. Many computation-oriented proposals are based on \citet{toulmin_2003}'s theory of practical argument. They distinguish between two main components of arguments,  "\textit{conclusion}" (also called "\textit{claim}") and "\textit{fact}" (also called "\textit{justification}" or "\textit{premise}"). They usually try to identify relations between components and between arguments, aiming to create a full argument tree that accounts for the argumentative structure of a text. This kind of model has been applied to essays \cite{stab-gurevych-2014-identifying} or user-generated discourse \cite{habernal-gurevych-2017-argumentation}. It is very general, thus easily portable to different domains. At the same time, it is not very stable, since inter-annotator agreement is not high, and the information it provides about the argument is not as rich as in the case of domain-specific approaches.

Another approach to modelling argument in texts are schemes. Argument schemes are 
%short ``\textit{abstractions substantiating the inferential connection between premise(s) and conclusion in argumentative communication}"~\cite{annotatingArgumentSchemesVisser}. They are 
``\textit{patterns of informal reasoning}"~\cite{walton_reed_macagno_2008} that ``\textit{represent forms of argument that are widely used in everyday conversational argumentation}"~\cite{macagnoArgSchemes2018}.
Argument Schemes specify a pattern of reasoning and a set of critical questions oriented to test
the defeasibility conditions on the pattern. This pattern and critical questions provide very insightful, actionable information about the argument, which can be later used for applications like building a counter-argument.

Several authors have adapted Walton's schemes to specific purposes, even proposing alternatives to critical questions~\cite{ATKINSON20181,kokciyan2018}. The main drawback of these proposals is that the inventory of scheme is very profligate,  
%\cite{LawrenceEtAl} propose to build a decision tree heuristic with yes or no questions to aid the annotator, but these trees can potentially become big and add a workload to the annotation process. 
and it has become clear that, identifying a scheme within a given text becomes quite difficult, both manually and automatically.%SHORT Later approaches have therefore tried to simplify the categories of analysis, while retaining as much as possible of its expressivity.

% We believe that critical questions can be generalized to a specific domain instead of a scheme based on the particular characteristics of it. Online Hate Speech, for example, usually refers to a collective and tries to associate a negative property, quality, consecuence or action to it, so a general strategy to counter it might be to question the relation between that collective and that property. But this generalization relaxes the most important properties of critical questions: that they always represent a defeasible condition of the argument. Indeed, there might be, for example, a tweet that doesn't explicitly mention a negative property associated to a collective, so the strategy proposed might not always be possible. It may also happen that a specific strategy don't produce the best quality response for a particular subset of hate tweets. Our hypothesis is that this problem can be overcomed by the diversity and multiplicity of strategies adopted. We believe that given multiple types of counter-narratives it is possible to select the most appropiate using evaluation or ranking systems. Variety of strategies also allows to adapt generated responses to a particular desired criteria or focusing on a particular aspect of the hate messages.\todo[inline]{Consultar con filósofos}

\subsection{The Periodic Table of Arguments}

Trying to find a trade-off between the excessive detail of schemes and the scarce information provided by claim-premise approaches, 
\citet{Wagemans2016ConstructingAP} proposes an analytic approach to argument schemes, aimed to obtain the core schemes proposed by~\citet{walton_reed_macagno_2008}, with fewer categories based on a limited set of general argument features.

This is a characteristic that we find particularly useful for building a simple system that is easy to annotate without an enormous effort and achieving a high level of agreement between human annotators, which leads us to expect higher reproducibility in inferred models. Moreover, an analytic approach allows determining which aspects of argumentation are more feasible to detect automatically, and identifying which particular aspects are more useful for a given application, such as components that could be used to elaborate a response.

All arguments under Wagemans's system have a premise and a conclusion labeled with one Type of statement each. But it goes beyond the mere premise-conclusion information. The PTA is a factorial typology of arguments that offers a comprehensive overview of the various types of arguments by describing them as a unique combination of three basic characteristics \cite{fourArgumentForms2019}: 

\begin{enumerate}
    \item \textbf{first order or second order} argument. A common term between premise and conclusion transfers the acceptability from one to the other. If this common term is explicit, %and no reconstruction of the argument is necessary to find it, 
    then it is a first order argument. If a reconstruction is needed, %SHORT (usually by transforming a premise into the subject of another premise with the predicate ``is true")
    then it is a second order argument.
    \item \textbf{predicate or subject} argument. If the common term is in the subject of the propositions making the premise and the conclusion, then it is a subject argument, otherwise, it is a predicate argument. 
    \item \textbf{policy, fact or value}. The conclusion and premise can be labeled each as a statement of policy (the speaker mandates or states that something should be done), a statement of value (the speaker issues an opinion about something), or a statement of fact (the speaker conveys a proposition as a true fact).
\end{enumerate}

\citet{annotatingArgumentSchemesVisser} conducted an exhaustive research on annotating the US 2016 presidential debate corpus using both Walton's schemes and Wagemans's Periodic Table of Arguments.
%SHORT and suggested revisions to the guidelines for annotating schemes that improved Cohen's $\kappa$. Though they managed to get an acceptable score for Walton's schemes, 
They reported a higher agreement for Wagemans's typology, specially without considering classification between first and second order arguments. Moreover, they sustain that for Wagemans's typology, ``\textit{the division into independent sub-tasks simplifies the annotation while maintaining reliability}".

%We adapted Wagemans's proposal to hate speech on social %media. Our goal is not to classify types of arguments, but %to identify elements that can be relevant to either a human %or a machine in the task of fighting the spread of hate.

We adapted Wagemans's proposal to hate speech on social media, with the goal of identifying elements that can be relevant to either a human or a machine in the task of analyzing or countering hate speech. 

Focusing on hate speech on Twitter, we have to take into account that
% in a corpus obtained from Twitter %"\textit{tends to contain a substantial amount of noisy data, which include spelling and grammar mistakes and innovations}" \cite{Schaefer2021ArgumentMO}. 
many argumentative hate tweets are based on assumptions justified by prejudice or context information that is difficult to recover. This means that in many cases, it is difficult to rebut them from the perspective of formal deductive logic. We believe that an approach based on informal logic, like the one proposed by \citet{Wagemans2016ConstructingAP}, is more adequate to capture this kind of arguments that are organized with informal relations.

In the following Section we describe our approach. We provide an overview of other social media corpora annotated with argumentative information in Appendix \ref{sec:corpora-social-media}. 

\section{A Framework to Identify Argument Components on Twitter Hate Speech}\label{sec:framework}

The goal of our argumentation model is to provide an argumentative analysis %meaningful information % with argument components that
%can be used to generate counter-narratives for hate speech tweets, both at annotation time by suggesting the annotator how to write the counter-narratives based on the previously labeled components and at prediction time using machine learning. %SHORT This is why all annotations are done considering how they might help in the task of generating counter-narratives, and they should be revised if at the moment of writing them, an alternative helps to build a better one.
that can help expose the core of the reasoning supporting a hate message. We believe that this can help both humans and automatic models to better address  hate speech. 

We are labeling two kinds of information: domain-specific and argumentative-general. Domain-specific information allows to exploit particular characteristics of hate messages on Twitter: they always mention a collective that is implicitly or explicitly associated with a negative property, action or consequence. Argumentative-general structure is based on a simplification of Wageman's proposal that is aimed to increase inter-annotator agreement. Reaching acceptable levels of inter-annotation agreement is very important to our purpose, as it indicates that the annotation process can be systematized and possibly automatized.

%Overall, annotation must be simple and clear enough that it can be reproduced by other humans with minimum training. 

%\todo{Ver si de acá no nos pueden criticar que sea una propuesta ad hoc o algo así, falta agregar más justificación? En definitiva lo que nosotros queremos anotar no es ni los componentes argumentativos por separado ni las contra-narrativas solas sino un conjunto de ambas que se complementen bien y produzcan resultados con sentido que puedan ser aprendidos por una máquina.}

%The US 2016 presidential debate corpus is a structured debate with a constrained set of speakers communicating in one particular genre. Visser et al. address the problems that arise when the annotators have to deal with interpreting propositions that are incomplete or not syntactically well-formed. This occurs sporadically because of bad transcriptions or interruptions during the debate. 

We created an annotation protocol\footnote{Annotation guidelines can be found at \url{shorturl.at/cv458}.} where both kinds of argumentative information are defined in a procedural manner. This protocol was applied by human analysts to annotate hate speech tweets, with five steps that are described as follows. The annotation team and environment are described in Appendix \ref{app:annotation-team-and-environment}.

\begin{figure*}
    \vspace{-0.4cm}
    \centering
    \includegraphics[width=\linewidth,trim={0 2.5cm 0 0.5cm} , clip]{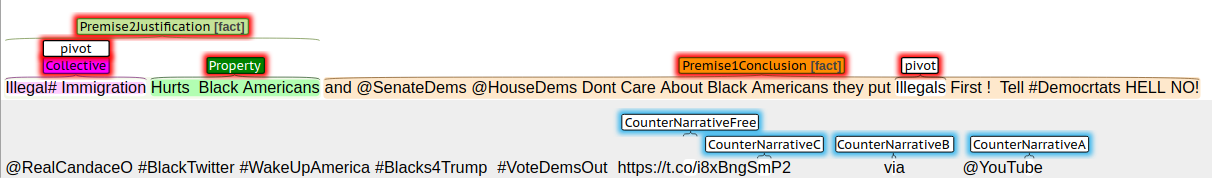}
    \vspace{-0.3cm}
    \caption{\label{fig:argumentative} Example of labeled argumentative hate tweet.%, with marked Justification and Conclusion (of type \textit{fact}), Collective, Property and Pivot.
    }
    \vspace{-0.3cm}
    %\vspace{-0.5cm}
\end{figure*}
% \todo{add more examples}

\subsection{Argumentative or Non-argumentative}
%SHORT We consider a tweet to be argumentative if there is ``\textit{a statement that is put forward in order to support another statement whenever the acceptability of the latter is in doubt}"~\cite{fourArgumentForms2019}.
%SHORT In order to do this, we need to consider relationships among different segments inside each of them. This is why the annotation phases of segmenting a tweet and classification between argumentative and non-argumentative are considered as a single phase: 
Following~\cite{fourArgumentForms2019}, ``\textit{an argument (...) consist(s) of two statements, namely a conclusion – the statement that is doubted – and a premise}". We consider a tweet to be argumentative if it is possible to divide it in these two components. 
% \todo{Pasar lo que sigue de arg/no-arg al apendice incluidos los ejemplos. Pasar figura 2 y figura 3 y agregar más ejemplos}
Examples of non-argumentative tweets can be found in Appendix~\ref{app:examples}: exhortations to some action without justification, insults, name callings, support for a particular policy or description of facts without an explicit conclusion. 
%All tweets of our dataset are classified either as argumentative or as non-argumentative. If the tweet is non-argumentative, no further annotation is made % and no counter-narratives are written. 
%(see examples in Figure~\ref{example:non-arg-tweet} in Appendix~\ref{app:examples}).

%Figure \ref{fig:non-argumentative} shows an example of a non-argumentative tweet.

% \begin{figure*}
%     \centering
%     \includegraphics[width=.8\linewidth]{example-non-argumentative-tweet-2.png}
%     \caption{\label{fig:non-argumentative} Example of non-argumentative hate tweet. The tweet displays a xenophobic slogan without explanation or justification}
% \end{figure*}

\subsection{Domain-specific components: Collective and Properties}

All hate messages are directed towards a specific group %SHORT or community 
by definition.
%\todo{Nota Diego: Recomendamos que citen la definición o al autor que trabaja este concepto específico, dado que el concepto de grupo puede diferir del de comunidad.}
Usually, the content of the message is to associate this group with a negative property or an undesirable action or consequence. If this property\footnote{A property is anything that is associated with the targeted community, whether it is an adjective, a consequence, an action, etc.} is explicit, we label it.

%SHORT This labelling can also be used for a separate task defined as automatically identifying what is being said about a targeted community.

\subsection{General argumentative components: Justification and Conclusion}

%If the tweet is argumentative, this means that it is possible to identify a Conclusion and a Justification. 
All argumentative tweets are labeled with one and only one
Conclusion and Justification, though these can be separated in many non-contiguous parts inside the tweet. Annotators were instructed to choose the longest Conclusion and Justification that they could find, %try to fit the most of the example inside these two components, 
leaving out only hashtags indicating topics, links, user mentions or non-relevant words or information. Justifications may be arguments themselves, having their own inner structure involving different premises, but this is not annotated as we are only interested in capturing the main standpoint that the user wants to gain acceptability for.

%In this classification is based on a simplified version of Wagemans's Periodic Table of Arguments, but without 
When labeling these components, we are not considering the subject-predicate structure proposed by Wagemans. \citet{annotatingArgumentSchemesVisser} warned about how this model presupposes that premises and conclusions of arguments consist of complete categorical propositions comprising a clear subject-predicate structure, which is not always the case in user-generated, informal social media text.
%SHORT and the problems that arise in the US presidential debate corpus as a result of ``interruptions, corrections, and general obscurity" that causes the transcription of the speech to be ``incomplete or not syntactically well-formed". 
%We detected the same problem to be even more severe in social media noisy user generated examples, so our proposal aims to simplify their identification.%SHORT Moreover, here we are not interested on classifying the arguments but rather on identifying the basic argumentative characteristics of the message.

\subsection{Argumentative relation: Pivot}

%\todo{If need for short, this could be gone}
%\citet{Wagemans2016ConstructingAP} classifies arguments as subject or predicate if their premises have a common element in the subject or in the predicate and as second order or first order if they need a reconstruction using an auxiliary sentence like "is true" (T) or not, respectively. We found that this classification is difficult since the common element is not always part of the same syntactic component and even when it is, it can fulfill multiple syntactic roles, so it is not easy to establish a syntactic correspondence between justification and conclusion.% This may be the reason why Visser et al notices that the classification of arguments as first-order or second-order leads to the least reliable results\cite{annotatingArgumentSchemesVisser}.

Following Wagemans, argumentative components transfer reason from one to the other. We assume that there can be textual cues of this transfer, in the form of an element that is common to both components. We call this element the pivot. We identify pivots as two sequences of words, one for each premise, that can be related to the element that those premises have in common. 

This relation is generally not unique; the underlying common ground between the premises could be expressed in different forms or could present multiple aspects signaled by different words. Whenever this element is explicit in the text (it might be not), we annotate it.

The pivot holds a relation with Wagemans's categories of first and second order arguments. If an argument is considered first-order, it means that the common element between premises must be explicit (by definition, it must be either of the form $A$ is $X$ because $A$ is $Y$ or $B$ is $Z$ because $C$ is $Z$). For a second-order argument, there might still be an explicit pivot or not.

\subsection{Types of Proposition}

Wagemans proposes ``a characterization of the types of arguments based on the combination of the types of propositions they instantiate'' \cite{Wagemans2016ConstructingAP}. These types are taken from debate theory \cite{SchutWagemans}, %\todo{Poner una cita aca. La cita que propone Wagemans para justificar esto es un libro en holandes que es imposible de leer para mi. Estaria bueno buscar algo que hable de estos tres tipos desde el punto de vista de teoría del debate que este en inglés} 
where three distinctions on propositions are made: (1) policy (P), (2) value (V) and (3) fact (F). 

We label our propositions using the same types and add to our annotation manual different guidelines on how to recognize each one: a policy proposition is a mandate often expressed as orders, imperatives, or actions that need to be accomplished in the public domain. Fact and value propositions were reported to be more difficult to differentiate. As a general criterion, a proposition to be labeled as value must have explicit markers of the speaker being involved in the assertion expressed (opinionated adjectives, verbs of thought, etc.). Otherwise, the premise is considered as fact.
%SHORT We found many examples where users express opinions as facts. Though many readers would recognize it as an opinion, if a truth value can be assigned to the assertion we label it as a fact, because they can be rebutted by asking  for sources or data to back the claim.
Examples can be seen %on Figure~\ref{example:factopinion} 
in Appendix \ref{app:examples}.
% \vspace{-0.03cm}
%\todo{If need for short, this could be out}We found many cases where opinions are expressed as facts, without explicit markers of the speaker involvement. Example \ref{example:factopinion} shows a premise labeled as "fact" where the user express her opinion as if it were a fact. We decided to label these examples as "facts" because, even though it can be obvious for many users that the message is expressing an opinion, it is still a good reply to make this explicit by asking sources or data to back the expressed claim. %SHORT A reason to justify this criterion is based on how this types of proposition are used to build counter-narratives: sometimes, even though it can be obvious for many users that the message is expressing an opinion, some may not notice and/or it is a good reply to make this explicit.
%\todo{if need for short this could be gone. Otherwise, check format: Examples, definitions, theorems, corollaries and similar must
%be written in their own paragraph. The paragraph must be
% separated by at least 2pt and no more than 5pt from the pre-
% ceding and succeeding paragraphs. They must begin with the
% kind of item written in 10pt bold font followed by their num-
% ber (e.g.: Theorem 1), optionally followed by a title/summary
% between parentheses in non-bold font and ended with a pe-
% riod. After that the main body of the item follows, written in
% 10 pt italics font (see below for examples).}

\section{The ASOHMO Corpus}\label{sec:corpus}

We applied our argumentation model via the annotation protocol described in the previous Section to the HatEval 2019 corpus~\cite{basile-etal-2019-semeval}. %13,000 tweets in English and 6,600 tweets in Spanish). 
Focusing on argumentative tweets, we did not annotate tweets labeled as ``aggressive", consisting mostly of abusive language (name callings, insults, exhortations to action and other types of attacks), nor tweets targeted against specific individuals or women, as they were almost exclusively abusive and non-argumentative. 
%SHORT We also filtered all tweets against women, since we noticed that the majority were also abusive and non-argumentative. 
After these filters, a corpus of 970 tweets in English and 196 tweets in Spanish remained.%SHORT, with a total of 25413 words in English and 5436 in Spanish and a significant proportion of argumentative tweets. Note that the HatEval tweets that were filtered out were implicitly labeled as non-argumentative. 

% Table~\ref{tab:nonargstats} on Appendix~\ref{app:statistics} shows the percentage of Non-Argumentative tweets per dataset and the percentage of argumentative tweets that are labeled with Collective and Properties, and with Pivots.
% Table \ref{tab:typeofpremisestats} on section \ref{app:statistics} shows the percentage of Justifications and Conclusions labeled as Facts, Policies and Values.
The dataset is released\footnote{
\url{https://github.com/ASOHMO/ASOHMO-Dataset}
} for the free use of the scientific community, together with the scripts for reproducing experiments.

\subsection{Inter-annotator Agreement}

\begin{table*}[ht!]\centering
    \begin{tabular}{l|c|cc|c|cc|cc}
        \hline
                         & & \multicolumn{2}{c|}{Domain-specific} & \multicolumn{5}{c}{Argument-general}\\
                          &   Argumentative  & Collective   & Property          &    Pivot       &   Justif.         &  Concl. & Type of Conc. & Type of Just.\\
        \hline

         $\kappa$    &  .85 & .64 & .60 & .52 & .62 & .64 & .60 & -.03\\
        \hline
    \end{tabular}
    \vspace{-0.3cm}
    \caption{Agreement scores between two annotators for 150 tweets.}
    \label{tab:agreement}
    \vspace{-0.5cm}
\end{table*}

% \begin{figure*}
%     \centering
%     \includegraphics[width=.9\linewidth]{agreements.png}
%     \caption{\label{fig:agreements} Cohen's $\kappa$ and F1 values for the different categories being annotated}
% \end{figure*}

%\todo{Poner con o sin overlap. Calcular el F1 de la clase minoritaria}
We calculated inter-annotator agreement to assess the reproducibility of the annotations and the feasibility of automatic identification. %The dataset was labeled by two annotators in a three stage process. After each stage, they updated the manual based on the difficulties found. 
%The first annotator annotated 800 tweets in English and 196 in Spanish, while the second annotated 170 tweets in English.
While the whole corpus was annotated by a single annotator, 150 tweets (15\% of the corpus) were labeled by a second annotator\footnote{The sample's size for the test is proportionally higher than many of the previous works: \citet{bosc-etal-2016-dart} used 100 tweets to calculate agreement over a dataset of 4000 whereas \citet{dusmanu-etal-2017-argument} used 100 tweets for its first dataset of 1887 tweets, 80 tweets for its second dataset of 1459 tweets and used the whole third dataset of 368 tweets.}. Then, per-category agreement was calculated with Cohen's $\kappa$ \cite{cohen1960}. Agreement was calculated in a per-tweet basis for the Argumentative vs. Non-Argumentative category using a binary label, and for the Type of Conclusion and Justification categories, using one label with three possible values representing \textit{fact}, \textit{value} and \textit{policy}. For all other categories, agreement was calculated in a per-word basis with a binary label assigned to each word, marking whether it belongs to the category or not.

% \todo{explicar mejor esto}
%We also report F1 score for this annotator against the tweets in the published dataset, and the best performing automatic models for each category. We want to assess the ability of automatic models to mimic a human annotator, compared to other human.

In Table~\ref{tab:agreement} we can see that annotators can reach a substantial level of reproducibility, around $\kappa=.6$ for Collective, Property, Justification, Conclusion and Type of Conclusion and $.85$ for the distinction between Argumentative or non-Argumentative tweets. In contrast, the Pivot presents a moderate level of inter-annotator agreement, and the Type of Justification presents no agreement at all.
% \todo{Agregamos ejemplos de disagreement? IDEA: poner un archivo en github con ejemplos y poner el link a ese archivo? Sino no va a entrar}
%Some examples of disagreement between annotators can be seen in Annex~\ref{app:examples}.
%\todo{calculate overlapping agreement}

%\todo{Add this after agreement calculation with overlap is calculated}
To calculate agreement, we follow a criterion similar to that of \newcite{annotatingArgumentSchemesVisser}: while comparing two annotators, if at least 50\% of the words in the smallest component marked by one of the annotators overlaps with words marked by the other one, then it is considered an agreement. For example, if one annotator marked "\textit{the damage illegals do}" as a Property associated to a Collective and the other annotator marked just "\textit{damage}" as a Property we consider that 100\% of the words in the shortest "\textit{damage}" in both examples and assume that all the other words are marked as not being part of the Property in both cases.

%SHORT Annotators reported three main difficulties while annotating the corpus. The first one is differentiating fact premises from value premises. The second one is labelling examples that might have both a "factual" style and exhortations towards an action. The third one is identifying the pivot. Annotators also reported a minor difficulty, once the two main premises are found, on classifying them as justification or conclusion. We found some cases where annotators could exchange the labels of justification and conclusion, and notice that annotators sometimes agreed on identifying the parts of a tweet but disagreed on which was the justification and which was the conclusion.

% The disagreement over the Type of Justification seems to be due to the fact that some \textit{values} are very hardly distinguishable from \textit{facts}. %SHORT Figure~\ref{fig:disagreement_type_justification} shows and discusses an example illustrating this difficulty.
% \begin{figure*}
%     \centering
%     \includegraphics[width=\linewidth]{disagreement_386_justification_type_dami.png}\\
%     \includegraphics[width=\linewidth]{disagreement_386_justification_type_jose.png}
%     \caption{Disagreement between annotators concerning the type of justification. The annotator above believes the justification is a fact, while the annotator below considers the speaker is stating an opinion. Note that there is also disagreement concerning the pivot, the collective and the property.}
%     \label{fig:disagreement_type_justification}
% \end{figure*}
\begin{figure}[H]
    \flushleft
%    \texttt{Sanctuary Cities are against the Law. Please shut them down & arrest/prosecute all criminal Governors & Mayors.}
    \small
    \texttt{@user @user \textit{\underline{sanctuary cities} are \textbf{against the law}.}\textsc{please shut \underline{them} down \& arrest/prosecute all \textbf{criminal} governors \& mayors}}\\
    \vspace{-0.2cm}
    % \texttt{@user @user \textit{sanctuary cities are \underline{against the law}.}\textsc{please shut them down \& arrest/prosecute all \underline{criminal} governors \& mayors}}
    % \includegraphics[width=.85\linewidth]{disagreement_389_jose_pivot.png}
    % \vspace{-0.3cm}
    \caption{\small Disagreement concerning Pivot. One annotator is underlined, while the other is bolded. Justification is marked with italics and Conclusion with capitalization}
    \label{fig:disagreement_pivot}
\end{figure}

When inspecting examples of disagreement between annotators for Pivot, as shown in Figure \ref{fig:disagreement_pivot}, we found that in many cases both annotations could be considered accurate, %. The relation the pivot intends to capture between Justification and Conclusion is not necessarily logical or formal, but informal, and 
as there may be more than one possibility for annotators to tag. Furthermore, as the relation is very deep in the layers of meaning, annotators may interpret it as signalled by different surface features, and as a consequence they may tag different sequences of words while considering the same relation.

Finding patterns in the disagreements between annotators can be used to redefine categories \cite{teruel-etal-2018-increasing}. In a second annotation phase, we will be redefining the Pivot category to obtain more agreement between annotators. We understand that this element is particularly challenging, because it signals a very deep relation and its correspondence with  surface textual phenomena may not be direct, or multiple. That is why we plan to rethink it as a binary classification problem, where human judges are presented with one or more possibilities of Pivots for a given argument, and they have to say whether they consider any of them to be a valid Pivot for the example. 

%Lastly, it can be seen that not all categories that can be systematically identified by humans can be successfully identified by automatic annotators. Indeed, the best automatic annotator presents low performance for the distinction between Argumentative and Non-Argumentative, which is the category with highest inter-annotator agreement. In contrast, for Type of Justification, performance is better than human agreement, showing that the model could identify some systematicities that humans were not applying in their decisions. We will be studying the behavior of the models to try to capture such systematicities and integrate them to improve annotation guidelines. For Pivot and Property, with acceptable inter-annotator agreement, the performance of the automatic annotators was also worse than for other elements. Particularly for Pivot, the performance of the automatic model was also significantly worse than a human annotator.
%\todo{Colapse value and fact categories and calculate agreement again}

\section{Automatic Identification of Arguments}\label{sec:automatic}

We conducted several experiments to assess the feasibility that LLMs can automatically identify different argumentative aspects. %in monolingual (English) and multilingual (English and Spanish) settings. We did not conduct experiments using only the Spanish portion of the dataset because of its small size.

For each set of hyperparameters used, we fine-tuned the same language models using different random tweets for each partition, always respecting this proportion. We report the average of these three fine-tuned models' F1, Precision and Recall to detect or classify argument components. For multi-label classification, the macro average is calculated, otherwise, we report the score of the target class. We also report per-class F1 scores for the three possible Types of premises: Fact, Value and Policy.

%like predicting the type of premise or joint predictions of multiple information aspects, we report the macro averages between the possible labels and the non-averaged F1 scores of each class. For all other binary classification tasks, we report the F1 for the target class.%, that is, F1 of the words labeled as Property, Justification, etc.

%We only conducted experiments for the English portion of the dataset, because the Spanish portion is too small for training
%\todo{Hacer experimentos multilinguales y cambiar esto introduciendo los experimentos multilinguales}

%SHORT\subsection{Preprocessing and Models}
\vspace{-0.15cm}
\paragraph{Models.} We fine-tuned the following LLMs:

%Preprocessing is very important when dealing with tweets, since they tend to have lots of non-alphanumeric characters, user handles (@user-name), hashtags, emojis, misspellings, and other non-canonical text.
% Following \cite{nguyen-etal-2020-bertweet} and \cite{polignano2019alberto} we used a soft normalization strategy consisting of: (1) Character repetitions are limited to a max of three; (2) Tweet is lower cased; (3) User handles are converted to a special token \verb|@user| (or \verb|@usuario| for spanish); (4) Hashtags are replaced by a special token \verb|hashtag| followed by the hashtag text; (5) Emojis are replaced by their text representation using \emph{emoji} library\footnote{\url{https://github.com/carpedm20/emoji/}}, surrounded by a special token \verb|emoji|.

% If a label was assigned to part of a word, that word was split into potentially many words, to make each word have only one label. The published dataset is already preprocessed and ready to reproduce experiments.

% The following models were used for evaluation:

%\subsubsection{Logistic Regression}

%As a baseline, we used Scikit-learn implementation of Logistic Regression~\cite{scikit-learn} and modeled the input data with two different approaches: using a bag-of-words and using contextual embeddings generated by extracting the last layer of the RoBERTa model we also evaluated. We tried four different values for inverse of regularization strength: 1.0, 0.1, 0.2 and 0.5. 

\paragraph{RoBERTa}\cite{liu2019Roberta}: a BERT-like \cite{devlin2018bert} LLM, pre-trained with more data. %, different hyper-parameters, larger learning rates and mini-batch sizes and removing the next-sentence prediction pretraining objective. When released in 2019, RoBERTa was established as new state of the art for 4 out of 9 GLUE tasks and matched state-of-the-art on other 2.

% \subsubsection{BiLSTM-CNN-CRF}

% We used the BiLSTM-CNN-CRF implementations by UKP Lab~\cite{reimers} for sequence tagging, composed of a BiLSTM with a Conditional Random Field linear classification layer at the end and char embeddings obtained by a 1D-CNN used to enrich the input. Besides the standard configuration, we also tried switching the CRF layer for a Softmax and using three different word embeddings as input for the model: FastText \cite{bojanowski2016enriching}, Komninos \cite{komninos-2016} and contextual embeddings generated using ELMO~\cite{elmo}.

\paragraph{BERTweet}\cite{nguyen-etal-2020-bertweet}: a RoBERTa-based LLM trained on data from Twitter.% domain-specific pre-trained language model trained using the RoBERTa procedure with more than 850M tweets. On specific tweet NLP tasks%SHORT like Part-of-speech tagging, name entity recognition and text classification
%, it has been shown that BERTweet outperforms baselines like RoBERTa and XLM-R.%SHORT A Twitter domain specific model like this one should be able to better capture the more colloquial style of Twitter full of slangs and informal expresions.

\paragraph{XLM-Roberta} \cite{xlm-roberta}: a RoBERTa based multilingual LLM. % masked language model based on Roberta training protocol but with 100 different languages. 
We fine-tuned it with a Mixed Language (Mix) version using both English and Spanish for training and testing and with a Cross-Lingual version (XL) using English for training and Spanish for testing.% We used this model as base for two experiments, using both languages for training and testing and using English for training and Spanish for testing.
%\subsection{RoBERTuito}

%RoBERTuito \cite{perez2021robertuito} is also a Twitter domain-specific pre-trained language model similar to BERTweet but trained for Spanish with more than 500M tweets. Given the characteristics of the dataset used for its training, where language was not restricted only to Spanish and almost 25M tweets were detected as written in English, Robertuito, besides performing very well for the Twitter domain, also possess some multilingual features.

%The following subsections describe the different experiments we performed.
\subsection{Predicting Individual Components}\label{ref:single-predictions}

We trained different kinds of models to automatically recreate the annotation process one component at the time: one for sequence binary classification, to predict if a tweet is argumentative or not; five models for token classification, to predict for each word, if it is labeled as part of the collective, the property associated to that collective, the pivot, the justification or the conclusion, respectively; and two models for sequence classification, fed only with the correspondent text of the premise (Justification or Conclusion), to predict the Type associated with it (fact, value or policy). Results of this experiment are shown in table \ref{tab:single-predictions}.

%\todo{Use confusion matrix to explain why type of justification has such a low performance}
\begin{table*}[ht!]
   \vspace{-0.6cm}
   \centering
   \setlength\tabcolsep{4pt}
   \renewcommand{\arraystretch}{0.8}
\begin{tabular}{|l|ccc|ccc|ccc|ccc|} %SHORT ccc|
        \hline
                        & \multicolumn{3}{c|}{RoBERTa} & \multicolumn{3}{c|}{BERTweet} &
                        \multicolumn{3}{c|}{XLM-RoBERTa-Mix} & \multicolumn{3}{c|}{XLM-RoBERTa-XL} \\
        \hline
                        & F1 & Pr & Rec & F1 & Pr & Rec & F1 & Pr & Rec & F1 & Pr & Rec \\
        \hline
        Arg./Non-Arg.& \textbf{.89}\small{$\pm$.02} & .84 & .95 & .88\small{$\pm$.01} & .84 & .93 & .87\small{$\pm$.04} & .84 & .91 & .84\small{$\pm$.03} & .84 & .85\\
        \hline
        Justification & .73\small{$\pm$.05} & .69 & .76 & \textbf{.77}\small{$\pm$.05} & .75 & .78 & .76\small{$\pm$.05} & .71 & .81 & .75\small{$\pm$.01} & .71 & .80\\
        Conclusion & .55\small{$\pm$.02} & .60 & .51 & \textbf{.61}\small{$\pm$.03} & .64 & .58 & .60\small{$\pm$.02} & .59 & .61 & .54\small{$\pm$.03} & .59 & .49\\
        Type of Just. & .41\small{$\pm$.09} & .48 & .39 & \textbf{.42}\small{$\pm$.09} & .48 & .41 & .35\small{$\pm$.05} & .34 & .37 & .33\small{$\pm$.03} & .33 & .35\\
        Type of Conc. & .58\small{$\pm$.05} & .62 & .57 & \textbf{.65}\small{$\pm$.11} & .67 & .65 & .61\small{$\pm$.06} & .65 & .62 & .63\small{$\pm$.02} & .66 & .62\\
        Collective    & \textbf{.59}\small{$\pm$.03} & .56 & .64 & .58\small{$\pm$.05} & .55 & .62 & \textbf{.59}\small{$\pm$.06} & .58 & .60 & .27\small{$\pm$.07} & .41 & .21\\
        Property    & .46\small{$\pm$.04} & .52 & .41 & .47\small{$\pm$.03} & .50 & .43 & \textbf{.50}\small{$\pm$.03} & .57 & .43 & .42\small{$\pm$.04} & .42 & .43\\
        Pivot    & \textbf{.45}\small{$\pm$.04} & .52 & .41 & .40\small{$\pm$.08} & .43 & .39 & .39\small{$\pm$.08} & .42 & .38 & .33\small{$\pm$.08} & .41 & .27\\\hline
        \end{tabular}\vspace{-0.3cm}
    \caption{F1, precision and recall for the target class in the automatic detection of argument components in tweets. Each experiment was carried out with three randomized partitions, the mean and standard deviation of the F1 are presented. Best results for F1 for each category are highlighted in boldface.} %SHORT on test dataset using RoBERTa base, BERTweet and RoBERTuito. Bold values are per-model best results}
    \vspace{-0.2cm}
    \label{tab:single-predictions}
\end{table*}

\begin{figure}
% \begin{table*}[ht!]\centering
% \renewcommand{\arraystretch}{0.8}
%     \begin{tabular}{|l|ccc|cc|c|cc|}
%         \hline
%                           &   Collect.   & Prop.          &    Pivot       &   Justif.         &  Conc. & Arg. & Type Conc. & Type Just.\\
%         \hline
%         human annotator F1   &  .65 & .65 & .54 & .84 & .76 & .96 & .77 & .45\\
%         automatic annotator F1   &  .59 & .50 & .45 & .77 & .61 & .89 & .70 & .55\\
%         \hline
%     \end{tabular}
%     \vspace{-0.3cm}
\includegraphics[width=\linewidth]{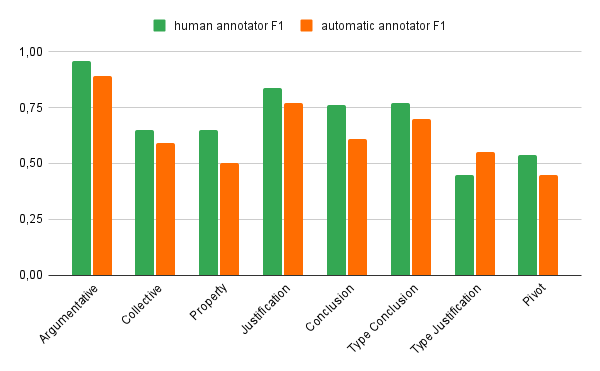}
    \caption{F1 score for ``predictions" done by a human annotator and compared with predictions done by the best performing automatic classifiers (BERTweet for Justification, Conclusion and their Types - trained with all the premises -, Roberta for Argumentative and Pivot and XLM-Roberta for Collective and Property).}
    \label{tab:human-f1}
\end{figure}
%\end{table*}

Distinguishing argumentative from non-argumentative tweets achieves a very satisfying .89 F1. In general, components with higher inter-annotator agreement perform better, with justifications identified with .77 F1. Components with low inter-annotator agreement are also identified with more errors: conclusions have an F1 of .61 ($\kappa=64$), collective F1=59, ($\kappa=64$), property F1=.50 ($\kappa=60$) and finally pivots only reach an F1 of .45 ($\kappa=52$).

In Figure ~\ref{tab:human-f1} we compared the F1 scores of the best performing models with inter-annotator agreement. We calculated the F1 score of the 150 tweets labeled by two judges using one as the ground truth and the other as the one being evaluated. We can see that both scores are highly correlated, although human annotators tend to agree slightly more than the automatic predictor with respect to the human ground truth, so there is still room for improvement for automatic predictors. % Table \ref{tab:human-f1} compares that score with the best performing automatic models.

Analyzing predictions for the worst performant components (Property and Pivot), we can see that the models predicting Properties have a tendency to recognize any word with a negative charge, disregarding if it is referring to the Collective itself. 
% Figure~\ref{example:property1} in section \ref{app:analyzis-of-differences} shows a tweet with no property originally labeled where predictor marked ``human trafficking".
Models recognizing Pivots sometimes find more than one possibility to label.
%and they tend to mark everything.
% \todo{add this examples in the paper, remove them from appendix}
Figure \ref{example:pivot1} shows how the model predicts the real pivot, but then also predicts another one not labeled on the original example that could be valid. This shows that, at least partially, some mistakes are made because of the subjective nature of the task and the multiple valid possibilities of labelling. To overcome this problem, annotators should consider the possibility of multiple pivots and try to label them all.

\begin{figure}[H]
    % \centering
    % \begin{quote}
    \vspace{-0.1cm}
    \small
        \texttt{\color{blue}\underline{\textbf{Salvini}} prosecuted for defending italian sovereignity and finally preventing hundreds of migrants to \textbf{invade} Italy \color{red}
        grande \underline{\textbf{Salvini}}, help us preserve the european culture against the \textbf{invasion} \#StopIslamization \#ComplicediSalvini \#StopInvasion \#RefugeesNotWelcome}\\
        
        % \textbf{Real Pivot}: Salvini - Salvini\\
        % \textbf{Predicted}: Salvini - Salvini \& invade - Invasion\\

    % \end{quote}
     \vspace{-0.4cm}
    \caption{Example of prediction of Pivot. Labeled justification is in blue, while conclusion is red. Real pivot is underlined, while predicted Pivot is bolded.}
    % \todo{ojo aca pareciera que esta mal, pero en el texto estamos diciendo que la alternativa es posible solo que no fue marcada. Revisar}
    \label{example:pivot1}
    \vspace{-0.3cm}
\end{figure}

Regarding the different models, BERTweet achieves the best performance on most experiments involving Justifications, Conclusions or their types, and is close to the best results on other components. Roberta achieves significantly higher results when identifying pivots. 

Multilingual experiments achieve similar performance than their monolingual counterparts for most components and the best performance for Properties, indicating that training with mixed languages do not decrease performance and can even improve it. 

Results on cross-lingual experiments where models are trained with English and tested against Spanish, on the other hand, show different behavior depending on the component: for finding argumentative tweets, Justifications, Types of Justification and Types of Conclusion, results are similar to their counterparts on monolingual experiments. Collective, on the other hand, has a major drop in performance for all experiments compared to all other model settings. This is explained because of the very specific lexicon used for naming collectives, with lots of out of vocabulary and slang words. The pivot also suffers a drop in performance on both multilingual settings, but specially on the cross-lingual one.

\subsection{Predicting Components Simultaneously}
\label{ref:joint-predictions}

% \subsubsection{Dependent aspects emulating the annotation process}
The goal in this case is to measure the performance of the models when simultaneously predicting components labeled on the same annotation step. We want to assess whether training with information about both components helps to improve the performance when predicting each of them individually or not.
We ran an experiment to jointly predict Collective and Property and another for Justifications and Conclusions. 
%SHORT In both cases, we filtered non-argumentative tweets. 
Each word is assigned one of three labels, indicating if they belong to either of the two searched components or not.

Joint prediction of components labeled on the same annotation step %(Table~\ref{tab:results-joint-predictions}) 
produces almost the same results as predicting them individually. This has the advantage of consuming half of the resources and time; however, the definition of the problem changes, as each token can only be part of one or none component, but not both.

\subsection{Predicting the Type of Premises}
\label{ref:type-of-premise}

\begin{table*}[ht!]
\centering
% \vspace{-0.1cm}
\setlength\tabcolsep{3.5pt}
\renewcommand{\arraystretch}{0.8}
\begin{tabular}{|l|cccc|cccc|cccc|cccc|} %SHORT ccc|
%         & \multicolumn{3}{c|}{LR w/embed} & \multicolumn{3}{c}{RoBERTa} \\
%        & F1 & Pr & Rec & F1 & Pr & Rec\\ %SHORT & F1 & Pr & Rec \\
\hline
                        & \multicolumn{4}{c|}{RoBERTa} & \multicolumn{4}{c|}{BERTweet} &
                        \multicolumn{4}{c|}{XLM-RoBERTa-Mix} & \multicolumn{4}{c|}{XLM-RoBERTa-XL} \\
\hline
                        & Macro & F & V & P & Macro & F & V & P & Macro & F & V & P & Macro & F & V & P \\
\hline

        \multicolumn{17}{|c|}{Models trained with both Justifications And Conclusions}\\
\hline        
%        Type Justif & .48\small{$\pm$.06} & .54 & .46 & \textbf{.64}\small{$\pm$.06} & .63 & .65 \\
        Type of Just & .49\small{$\pm$.07} & .92 & .13 & .41 & .53\small{$\pm$.08} & .93 & .19 & .47 & \textbf{.55}\small{$\pm$.01} & .94 & .37 & .34 & .52\small{$\pm$.17} & .93 & .41 & .21 \\ 
        Type of Conc & .63\small{$\pm$.14} & .82 & .22 & .85 & \textbf{.70}\small{$\pm$.14} & .85 & .37 & .87 & .67\small{$\pm$.16} & .78 & .37 & .86 & .57\small{$\pm$.04} & .78 & .34 & .60 \\
        Type of both & .66\small{$\pm$.05} & .90 & .28 & .79 & \textbf{.69}\small{$\pm$.12} & .91 & .34 & .82 & .67\small{$\pm$.04} & .88 & .35 & 79 & .60\small{$\pm$.03} & .89 & .39 & .53 \\
%        Type Conc & .60\small{$\pm$.03} & .60 & .60 & \textbf{.73}\small{$\pm$.03} & .75 & .71\\

        \hline
        \multicolumn{17}{|c|}{Models trained with just one of them}\\
\hline  
        Type of Just & .41\small{$\pm$.09} & .95 & .28 & .00 & \textbf{.42}\small{$\pm$.09} & .95 & .13 & .17 & .35\small{$\pm$.05} & .97 & .00 & .08 & .33\small{$\pm$.03} & .95 & .0 & .05 \\ 
        Type of Conc & .58\small{$\pm$.05} & .81 & .07 & .85 & \textbf{.65}\small{$\pm$.11} & .84 & .20 & .89 & .61\small{$\pm$.06} & .70 & .28 & .85 & .63\small{$\pm$.02} & .78 & .45 & .65\\
        \hline
    \end{tabular}
    %\vspace{-0.1cm}
    \vspace{-0.3cm}
    \caption{Results for identification of Type of premises tested against both Justification and Conclusion, only Justifications and only Conclusions. Results are compared against those achieved by the best performing model trained with only one of the two kinds of premises.}
    \label{tab:results-premise-types}
    \vspace{-0.3cm}
\end{table*}

%\paragraph{Predicting the type of premises}
%\subsubsection{Predict type of premise}

The Type of Conclusion or Justification (Fact, Policy or Value) should be independent of its premise (Justification or Conclusion), so in terms of semantic information, to predict this, it should not matter if models are trained with just one or both of them. 

Moreover, using both kinds of premises increases the number of training examples and can help to overcome the unbalance between Fact and Policies (specially on Justifications, where facts are the vast majority).% We assess (Table~\ref{tab:results-premise-types}) if such a model can achieve better performance than models trained only one kind of premise. The model is tested against the same datasets used to test models trained with only one kind of premise, and then with a combination of those two datasets.
In Table~\ref{tab:results-premise-types} we can see that models trained to predict the Type of Premise with both Justifications and Conclusions perform much better than models trained with just one or the other. For Type of Justification, these models achieve F1 scores that are between 10 and 20 points higher. For Type of Conclusions, their F1 scores are around 5 points higher. When checking the per-class F1 scores, the improvement in performance is concentrated on the minority classes. For Type of Justification, both Value and Policy classes improve highly, and for Type of Conclusion the most difference is on the Value class.%SHORT Testing against the dataset combining both kind of premises also achieves a very high F1 score, higher than using models trained with just one kind of premise.

\subsection{Impact of training dataset size}

We want to assess how much data is needed for the models to achieve an acceptable performance. For this purpose, we ran several experiments following the same settings as in~\ref{ref:single-predictions} but using smaller portions of the original datasets. Our goal is to measure the impact of having smaller datasets for each component and the relative gain of adding new examples, considering that the task of labeling them is expensive. We used a random sample of 25\%, 50\% and 75\% of the corpora used for training and compare the F1 scores with those obtained by the models trained with the whole corpus.

\begin{figure*}
    \centering
    \includegraphics[width=.50\linewidth, height=.37\linewidth, clip]{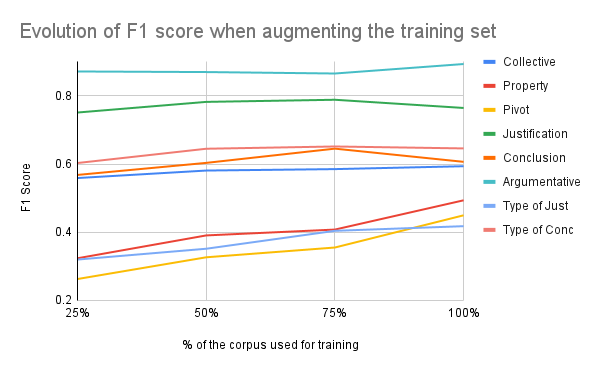}
    \hspace{-0.3cm}
    \includegraphics[width=.50\linewidth, height=.37\linewidth, clip]{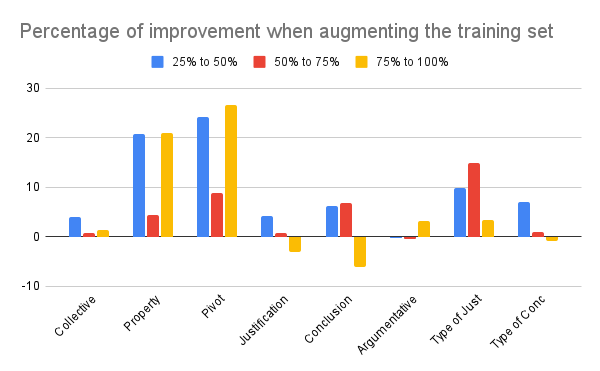}
    \vspace{-0.5cm}
    \caption{\label{fig:evolution-incremental} Evolution of F1 scores per argumentative component when increasing the size of the dataset used for training. The figure on the left shows F1 score absolute values, while the one in the right shows the percentage of the score wrt the final value obtained when reducing the dataset.%, with marked Justification and Conclusion (of type \textit{fact}), Collective, Property and Pivot.
    }
    %\vspace{-0.5cm}
\end{figure*}

Figure~\ref{fig:evolution-incremental} compares the F1 scores of the best performing models for each component in~\ref{ref:single-predictions} with those obtained by the same models trained with smaller portions of the same datasets. On the left, we show the evolution of the F1 score when increasing the size of the training dataset. On the right, we show the percentage of improvement of the F1 score between each size of the dataset for each component. For example, the model predicting Pivot trained with 75\% of the dataset achieved a score of 0.36 while the model trained with the whole dataset (100\%) achieved a score of 0.45, which represents an improvement of 26.6\% of this score.

When looking at performance of models trained with smaller fractions of the dataset (figure \ref{ref:single-predictions}) we can see that those components with better scores can achieve similar results using fewer data, while components with worse performance (Property, Pivot and Type of Justification) are much more sensible to the amount of examples on the dataset. This could be considered as an indicator that the size of the dataset is enough for most components but for these last three, if more examples were added to the dataset performance could improve.

\section{Discussion of results}\label{sec:discusion}

We have seen that some argumentative aspects of hate speech in tweets can be successfully identified by Large Language Models (LLMs), namely, whether a tweet is argumentative or not and Justifications, Conclusions and the Type of Conclusions. 

This kind of information may be useful to provide an argumentative analysis of tweets, possibly for argument retrieval. It is probably also useful to guide the (semi-)automatic generation of some counter-narratives, like those that are aimed to question the Justification or those aimed to some kinds of Conclusions, like Values or Policies. 

Domain-specific argument information, like Collective and Property, are not very successfully identified. Different strategies, like Named Entity Recognition approaches, may yield more satisfactory results.

Pivots, aiming to identify the relation between Justification and Conclusions, and a key component to reconstruct Wagemans's typology, cannot be successfully identified, either by humans or automatically. It seems that a different approach must be taken to identify them manually, possibly identifying all possible sequences of words that elicit a relation between Justification and Conclusion.

These results will be instrumental for the annotation of a bigger annotated corpus, specially for Spanish, and to integrate these concepts into LLMs.

\section{Summary and Future Work}

% \todo{

% %We found that models trained without information about other components usually predicts collectives when there is no explicit property written about them. The same thing happens when predicting Properties but less often.

% Some components are more sensible to language specific lexicons, like Collective.

% }

We have presented an approach to determine which aspects of argumentative information from hate speech in social media is liable to be integrated into LLMs. We have adapted the analytic approach of an informal logic based on \cite{Wagemans2016ConstructingAP} and have developed annotation guidelines which have then used to enrich a reference dataset for hate speech with argumentative information. 

%Our motivation is to provide an argumentative analysis that expose the reasoning behind hate messages in order to provide tools for humans or automatic models to counter its spread. This is the first work, to our knowledge, where Wagemans's argumentative theory is adapted to the context of hate messages on social networks.

%Our motivation is the belief that quality is preferred over quantity: having less tweets with extra information about the argumentative relations that are being attacked with each counter-narrative could lead to better results than just having a raw pair of hate speech -- counter-narrative. This is the first work, to our knowledge, where argumentative information is used for automatic counter-narratives generation and is also the first work, to our knowledge, of argumentative information annotated on noisy user-generated tweets.\todo{Esto era cierto cuando decía argumenation schemes en lugar de argumentative information como dice ahora. Sigue siendo cierto?}
We developed a robust annotation process and guidelines to obtain high agreement between annotators. Indeed, an initial assessment of inter-annotator agreement, shows agreement above $\kappa = .6$ for most categories, except the most interpretative ones. Considering we are dealing with user-generated text, we find this a very hopeful scenario. We are also working on adapting the categories with more disagreement, like Pivot, based on the patterns of the disagreemeent between annotators, so that in further annotation efforts they can be identified in a more reproducible ways, both by humans and automatic methods.

We show to which extent it is possible for Large Language Models to automatically identify the argumentative components, so that this kind of information can be integrated with purely data-driven approaches to enrich the analysis of text and produce more insightful, reasoned outputs.

%reproducing the annotation process in different ways and achieving good results in many of the proposed categories. We also show the performance when progressively incrementing the size of the training data as an indication that for many components, the amount of examples seems to be enough to achieve the best possible performance.
%SHORT while for others, expanding the dataset could lead to further improvements.%SHORT We compare the performance of automatic models with human annotators, highlighting the difficulties arising from the subjective nature of the task.
%With this purpose in mind, we took some of the components used by Wagemans to build his Periodic Table of Arguments and re-defined them not to be used as a classification system but as markers of simple argumentative relations between parts of an argumentative hate tweet that could expose possible weak point from where the tweet could be attacked. 
Finally, the published dataset is also a contribution to the existing corpora of argument mining on social networks. It is publicly available at \url{https://github.com/ASOHMO/ASOHMO-Dataset}.

%These corpora are scarce and as far as we are aware, this is also the first work that uses hate messages from Twitter. Considering that it is one of the Social Networks with the most amount of users and information generated all over the world and considering that it tends to be more noisy than other social networks, this represents a significant contribution to helping counter hate speech online.

For future work, we plan to annotate bigger corpora, focusing on improving reliability on difficult, yet potentially useful, components, like Pivot. We also plan to add counter-narratives associated to each tweet and train models to automatically generate them. We want to assess to which extent the argumentative information helps in better generating automatic responses.%SHORT for the hate tweets.

% \section{Ethical Considerations}

\section{Limitations and Ethical Considerations}

In the first place, we would like to make it clear that the human annotations presented here are the result of the subjectivity of the annotators. Although they have been instructed through a manual and training sessions, there are still significant variations between interpretations, and further researchers may assign different categories to the examples in the corpus.

Also, it is important to note that the automatic procedures obtained are prone to error, and should not be used blindly, but critically, with attention to possible mistakes.

Then, it is also important to note that the corpus used for this research is very small, specially in the Spanish part, so the results presented in this paper need to be considered indicative. A bigger sample should be obtained and annotated to obtain more statistically significant results.

The findings of this research can potentially inform the development and improvement of language models and chatbot systems. However, we emphasize the importance of responsible use and application of our findings. It is essential to ensure that the identified argumentative components are utilized in a manner that promotes reasoned usage and does not contribute to the spread of hate speech or harmful rhetoric. We encourage researchers and developers to consider the ethical implications and societal impact of incorporating argumentative analysis into their systems.

The data have been adequately anonymized by the original creators of the Hateval corpus.

Studying hate speech involves analyzing and processing content that may be offensive, harmful, or otherwise objectionable. We acknowledge the potential impact of working with such content and have taken steps to ensure the well-being of the research team involved. We have provided comprehensive guidelines and training to our annotators to mitigate any potential emotional distress or harm that may arise from exposure to hate speech. Additionally, we have implemented strict measures to prevent the dissemination or further propagation of hate speech during the research process.

Finally, we have not specifically conducted a study on biases within the corpus, the annotation or the automatic procedures inferred from it, nor on the LLMs that have been applied. We warn researchers using these tools and resources that they may find unchecked biases, and encourage further research in characterizing them.

\section*{Acknowledgments}

Annotation was done using the brat annotation tool \cite{stenetorp-etal-2012-brat}.

This work used computational resources from CCAD – Universidad Nacional de Córdoba (https://ccad.unc.edu.ar/), which are part of SNCAD – MinCyT, República Argentina.

% The acknowledgments should go immediately before the references. Do not number the acknowledgments section.
% \textbf{Do not include this section when submitting your paper for review.}

\bibliographystyle{acl_natbib}
\bibliography{acl2021}

\newpage
\newpage
\appendix

\paragraph{APPENDIX}
\section{Annotation team and environment}
\label{app:annotation-team-and-environment}

Two annotators (a philosopher and a computer scientist) have been trained with the guidelines described in section \ref{sec:framework}, with a three stage training process, where they labeled a first set of examples, discussed their difficulties, systematized further hints and criteria, updated the annotation manual and started again. We prioritized having the lesser amount of annotators doing the most possible amount of annotations. Our hypothesis is that the more annotators, the more difficult it is to reach a uniform criterion that can be understood in the same way by everyone. So fewer annotators doing more work should lead to more reliable annotations and to better inter-annotator agreement.

The average time for annotators to label a tweet is approximately 4 minutes per example. The annotation time changes depending on whether the tweet is argumentative or not. For argumentative tweets, the average time is around 5 minutes, while for non-argumentative tweets the average time is less than 1 minute.

The first annotator annotated 800 tweets in English and 196 in Spanish, while the second annotated 170 tweets in English.

\section{Corpus statistics}
\label{app:statistics}

Table \ref{tab:nonargstats} show the percentage of tweets that are labeled as non-argumentative in English and in Spanish, and also the percentage of tweets in each language that have a pair of Collective and Property and a Pivot labeled. Considering only the non-targeted and non-aggressive hate tweets against immigrants from HatEval, the majority of tweets are labelled as Argumentative in both languages. Regarding the Collective-Property pair and the pivot, the table shows the percentage of the final dataset that have them labeled. Table \ref{tab:typeofpremisestats} shows the percentage of Justifications and Conclusions that are labeled as \textbf{F}act, \textbf{P}olicy or \textbf{V}alue. Justifications have an ample majority of examples labeled as Fact, while the distribution between classes is more even when observing conclusions. In both cases, the "Value" class is the least frequent.
% \todo{agregar una minima descripcion con texto aca}

\begin{table}[h]\centering
    \begin{tabular}{|l|c|cc|}
        \hline
                          &   Non-Arg   & Collective          &  Pivot \\
        \hline

        English    &  25.3\% & 58.2\% & 45.1\% \\
        Spanish   &  26.5\% & 61.1\% & 37.5\% \\
        \hline
    \end{tabular}
    \vspace{-0.2cm}
    \caption{Percentage of tweets labeled as Non-Argumentative and with Collective-Property and Pivot labeled}
    \vspace{-0.4cm}
    \label{tab:nonargstats}
\end{table}

\begin{table}[h]\centering
    \begin{tabular}{|l|ccc|ccc|}
        \hline
                        & \multicolumn{3}{c|}{Justification}    & \multicolumn{3}{c|}{Conclusion}\\
        \hline
                          & F & P & V & F & P & V\\
        \hline

        English    &  93\% & 4\% & 3\% & 37\% & 57\% & 6\% \\
        Spanish   &  97\% & 2\% & 1\% & 56\% & 28\% & 16\% \\
        \hline
    \end{tabular}
    \vspace{-0.2cm}
    \caption{Percentage of Justifications and Conclusions labeled as \textbf{F}act, \textbf{P}olicy or \textbf{V}alue}
    \vspace{-0.5cm}
    \label{tab:typeofpremisestats}
\end{table}

\section{Preprocessing}

Preprocessing is very important when dealing with tweets, since they tend to have lots of non-alphanumeric characters, user handles (@user-name), hashtags, emojis, misspellings, and other non-canonical text. Following \cite{nguyen-etal-2020-bertweet} and \cite{polignano2019alberto} we used a soft normalization strategy consisting of:

\begin{itemize}
    \item Character repetitions are limited to a max of three
    \item User handles are converted to a special token \verb|@usuario|
    \item Hashtags are replaced by a special token \verb|hashtag| followed by the hashtag text and split into words if this is possible %laura would like to know more about this, how it is done
    \item Emojis are replaced by their text representation using \emph{emoji} library\footnote{\url{https://github.com/carpedm20/emoji/}}, surrounded by a special token \verb|emoji|.
\end{itemize}

\section{Experiment settings}

For all monolingual experiments we used 770 tweets of the English portion of the dataset as training (79\%), 100 tweets as development (10.5\%) and 100 tweets as test (10.5\%). Multilingual experiments were twofold: using both English and Spanish for both training and testing, and using English for training and development and Spanish for test. In the first case, we used 770 English and 120 Spanish tweets as training (76.3\% of the dataset), 100 English and 26 Spanish tweets as development (10.8\%) and 100 English and 50 Spanish tweets as test (12.9\%). In the second case, we used 850 English tweets as training (73\% of the total), 120 English tweets as development (10\%) and all the 196 Spanish tweets for testing (17\%).

In all cases, we tried 5 different values for learning rate (1e-05, 2e-05, 5e-05, 5e-04 and 5e-06) and used the development dataset to implement early stopping with a maximum of 10 epochs. Table \ref{tab:model-parameters} shows the values for the hyperparameters used on all models trained with our examples.

\begin{table}[h]\centering
    \begin{tabular}{|l|c|}
    
        \hline
                        Batch Size & 16\\
                        Optimizer & AdamW\\
                        Dropout & 0.1\\
                        Epochs & 10\\
                        Weight Decay & 0.01\\
                        Adam $\epsilon$ & 1e-06\\
                        Adam $\beta$1 & 0.9\\
                        Adam $\beta$2 & 0.99\\

        % English    &  93\% & 4\% & 3\% & 37\% & 57\% & 6\% \\
        % Spanish   &  97\% & 2\% & 1\% & 56\% & 28\% & 16\% \\
        \hline
    \end{tabular}
    \vspace{-0.2cm}
    \caption{Hyperparameters used for training all models used on our experiments}
    % \vspace{-0.5cm}
    \label{tab:model-parameters}
\end{table}

% \todo{Agergar una tabla con los hiperparametros y los LR seteados para cada experimento}

\section{Examples And Decisions From The Annotation Process}
\label{app:examples}

In the following section, we show examples of labeled tweets that illustrate particular decisions taken when defining the annotation protocol. Example \ref{example:non-arg-tweet} shows a frequent case of a non-aggressive, non-targeted and non-argumentative tweet, consisting on the expression of one or many stances or exhortation to one or many actions but without any explicit intention of connecting them.

Example \ref{example:factopinion} shows an example of a premise where a user states her opinion as a verifiable fact. Although it could be arguable that she is expressing a Value about a subject (immigration or assimilation), we consider all tweets that could be fact checked (specially if the user doesn't use explicit markers of her involvement in the statement) to be of type Fact.
Example \ref{example:collective-propery} shows an annotation of a Collective-Property pair. The Property is any negative concept, adjective, consequence or aspect of reality that is explicitly or implicitly associated with the target of the hate message. In this case, the tweet is stating that immigrants are not wanted by the people. The cases where there is no explicit association between Collective and Property are diverse, but we present three examples that we believe represent the majority of the cases. Example \ref{example:collective-property2} shows a case where instead of defining a negative property associated with the targeted collective, the user defines a positive Property associated with the absence of that Collective. Example \ref{example:collective-property3} shows a case where a negative Property is associated with the targeted Collective but in an indirect manner that must be reconstructed using contextual information not included in the hate message. In this case, the reference is made through the mention of "Operation SOAR", an operation made by the ICE in the United States specifically targeting immigrants registered as sex offenders and by the hashtag "StopTheInvasion" referring to a narrative built against immigrants as if there were a coordinated plan to invade a country. Example \ref{example:collective-property4} shows a case where the main standpoint of the tweet is an action that must be taken and there is no explicit mention of any Property. In these cases, the Property could potentially be reconstructed by appealing to find the motivation of these advertised actions, but it can not be labeled explicitly on the text of the tweet.

Example \ref{example:justification-conclusion} shows a Justification labeled as Fact and a Conclusion labeled as Policy. The main standpoint of the message must be first identified as Conclusion, and then any part of the tweet that fulfills the role of providing reasons for that standpoint is identified as Justification. One typical pattern frequent in many tweets is to express a mandate or policy that must be followed, usually with the form of a phrase or hashtag using the imperative mode, and a Fact (or less frequently also another mandate) that supports and aims to explain why that mandate must be followed.

Example \ref{example:pivot-ex} show a tweet with a Pivot. In this case, the user binds the "money" as a cause of immigration to conclude that "money" is not needed. All tweets present some aspect that links Justification with Conclusions, but not always that relation is mentioned directly. Example \ref{example:pivot-ex2} shows a tweet where no Pivot was labeled. The link between the premises relies on the implicit assumptions that the hate that they supposedly bring to the EU is against Christians and that because of that hate, Christians are not safe. But to recognize it, the relationship must be reconstructed using implicit information defined by the context, otherwise it is impossible to establish. In these cases, we do not label the Pivot for the sake of simplicity. We require that the relation between the two phrases constituting a Pivot is direct and easy to spot.

% \todo{agregar algo de texto aca}

\begin{figure}[H]
    % \centering
    % \begin{quote}
        \texttt{No to \#EU migrant camps in Libya, PM al-Serraj}
    % \end{quote}
    % \vspace{-0.2cm}
    \caption{Example of non-argumentative tweet}
    \label{example:non-arg-tweet}
    % \vspace{-0.7cm}
\end{figure}

\begin{figure}[H]
    % \centering
    % \begin{quote}
        \texttt{Anyone who, ACTIVELY OR PASSIVELY, subscribes to immigration and especially assimilation is joining the battle to destroy White}
    % \end{quote}
    % \vspace{-0.2cm}
    \caption{Premise of a tweet labeled as ``fact"}
    \label{example:factopinion}
        % \vspace{-0.7cm}
\end{figure}

\begin{figure}
    \texttt{@user Time to leave the uk commonwealth and Europe that would end immigration \textbf{people do not want more} \underline{refugees} enough is enough}
    \caption{Example of Collective and Property labeled. Collective is underlined while Property is bolded}
    \label{example:collective-propery}
\end{figure}

\begin{figure}
    \texttt{Good this makes it a safe country immigrants can now go home}
    \caption{Example of tweet without Collective and Property labeled. In this case, Property is associated with the absence of immigrants, therefore it is indirectly defined and not mentioned explicitly}
    \label{example:collective-property2}
\end{figure}

\begin{figure}
    \texttt{ICE officers arrest 32 sex offenders on Long Island as part of 'Operation SOAR' :link: \#StopTheInvasion \#SecureTheBorde}
    \caption{Example of tweet without Collective and Property labeled. In this case, the collective is not explicitly mentioned but referred through contextual information}
    \label{example:collective-property3}
\end{figure}

\begin{figure}
    \texttt{Canada is an immigrant country Don't change it to refugee country please}
    \caption{Example of tweet without Collective and Property labeled. In this case, the focus of the message is put into an action that must be taken and not on associating the Collective with a Property}
    \label{example:collective-property4}
\end{figure}

\begin{figure}
    \texttt{\color{blue}Victims of Illegal Alien Crime describe heartbreak, frustration \color{red}\#BuildTheWall \#ProtectAmerica \#EndChainMigration \#EndIllegalBirthrightCitizenship \#NeverForget the American Victims of Illegal Alien Migration}
    \caption{Example of labeling of Justification (in blue) and Conclusion (in red). Justification is labeled as Fact while Conclusion is labeled as Policy}
    \label{example:justification-conclusion}
\end{figure}

\begin{figure}
    \texttt{\color{blue}Why do foreign individual dump \textbf{money} (and refugees) into our country?  \color{red}We don't need their \textbf{money} and their programs.}
    \caption{Example of a tweet with a labeled Pivot. Justification is shown in blue while Conclusion is in red. Labeled Pivot is shown bolded}
    \label{example:pivot-ex}
\end{figure}

\begin{figure}
    \texttt{\color{blue}Nice tweet , Joyce, Truth is they flee Iran etc but want to bring their hate to the Eu \color{red}even in refugee camps Christians not safe.}
    \caption{Example of a tweet without a Pivot labeled. The Justification is shown in Blue while the Conclusion is in Red. The link between the two premises relies on the relation between "hate" and "not safe"}
    \label{example:pivot-ex2}
\end{figure}

\section{Disagreement between annotators}
\label{app:disagreement}

In the following section, we analyze examples of disagreement between annotators to better understand the aspects that are most difficult to systematize about annotating argumentative components.
Example \ref{agreement:collective-property} show a disagreement concerning Collective and Property. Here, one annotator didn't consider that there was a Collective and Property to label, while the other did. We found that most disagreements regarding these components are of this kind. If both annotators agree that the tweet has a Collective and Property to label, in most cases they agree also what parts of the text constitutes them. In the few cases where both annotators labeled a Collective and a Property, but they did not match exactly, they had a major overlap and only differed on adding a few more words at the beginning or at the end. Example \ref{agreement:collective-property2} shows a disagreement of such kind.
Example \ref{agreement:justification-conclusion} shows how both annotators agree on how to split the text but disagree on which part is the Justification and which is the Conclusion. To improve the annotation process, the guidelines should emphasize that the main standpoint of the tweet should be identified before labeling the Justification.
Example \ref{fig:disagreement_pivot} shows disagreement about labeling the pivot. In this case, each annotator found a different Pivot that could be considered correct. The annotation guidelines enforce each annotator to label only one Pivot but there are examples, like the one mentioned above, where multiple Pivots could be found. This indicates that there could be an opportunity of improving the system if we enforce annotators to label all possible Pivots.
Example \ref{agreement:type-of-policy} shows a disagreement on the Type of a Justification. The premise has declarative sentences with informative content (like "It is the third anniversary of her death") mixed with mandates or actions that must be followed ("Remember Kate Steinle today" and "We must not forget"). Depending on the part of the sentence

\begin{figure}
    \texttt{@user @user The idea is to bring in the \underline{"dreamers"} \textbf{so that they vote for Democrats} because Dems know they have to import their voters. That is literally the only reason the Democrats care about this issue. In the meantime, YES THEIR PARENTS}
    \caption{Example of disagreement concerning the Collective and Property. One annotator did not label any of them. Collective labeled by the other annotator is underlined while Property is bolded}
    \label{agreement:collective-property}
\end{figure}

\begin{figure}
    \texttt{\color{red}Mexico\color{black}'s not sending their best. \textbf{They are dumping their \underline{killers aka garbage} on us}. \#StopTheInvasion \#DeportThemAll \#NoAmnesty \#BuildTheWall}
    \caption{Disagreement concerning the Property. Collective labeled by both annotators is shown in red. Property labeled by one annotator is bolded while the one labeled by the other annotator is underlined}
    \label{agreement:collective-property2}
\end{figure}

% 295/510/536

\begin{figure}
    \texttt{\textbf{@user @user you come with the usual lies an insults.} \underline{Fact is that mass immigration} \underline{into Ireland has been going on} \underline{for decades, most illegal and} \underline{from other EU countries, still} \underline{trans-formative. All the people} \underline{
    seeking asylum}}\\
    \\
    \texttt{\underline{@user @user you come with the} \underline{usual lies an insults.} \textbf{Fact is that mass immigration into Ireland has been going on for decades, most illegal and from other EU countries, still trans-formative. All the people seeking asylum}}
    \caption{Disageement between annotators concerning Justification and Conclusion. Justification is bolded while Conclusion is underlined. While both annotators split the argument in the same fashion, they disagree on which part is the justification and which is the Conclusion}
    \label{agreement:justification-conclusion}
\end{figure}

\begin{figure}
    \texttt{\textbf{Remember Kate Steinle today. It is the third anniversary of her death We must not forget.} \#KateSteinle\#IllegalAliens
    \#OpenBorders\#BuildThatWall
    \#MondayMorning\#ImmigrationReform
    \#ImmigrationIsAWeapon}
    \caption{Example of disagreement concerning type of premise. Justification (bolded) was labeled as Fact by one annotator and as Policy by another}
    \label{agreement:type-of-policy}
\end{figure}

\section{Analysis of differences between automatic classifications and ground truth}
\label{app:analyzis-of-differences}

We analyze the errors made by automatic classifiers when recognizing argumentative components, trying to determine possibilities of improvement either in the annotation process or in the settings of the task of automatic recognition.

Example \ref{example:non-arg} show an example of a non-argumentative tweet that was classified as argumentative by the automatic predictor trained as described in \ref{ref:single-predictions}. The tweet has several hashtags calling for actions, but there is no explicit intention of using any of them as a justification of the others. The tweet refers to a mother who supposedly needs prayers, indicating that the author is aware of a context that is missing for us.

Example \ref{example:colective1} shows a prediction done by a model trained following the settings described in \ref{ref:single-predictions}. Here, the model correctly identifies a Collective mentioned in a xenophobe tweet, but there is no explicit Property assigned to them and because of this, it shouldn't have been labeled. Though this model was sometimes able to distinguish when the Collective should have been labeled or not, we found this error to be very frequent in experiments done with these settings. This led us to propose the experiment described in \ref{ref:mising-components} separating the problem in two: first identifying if there is a pair of Collective and Property to label and then finding them on the tweet. When scoping the problem to find a Collective in a tweet that we know it is present, most errors produced by the automatic classifiers are discrepancies on the amount of words used to refer to the collective (like in example \ref{example:collective2}) or whenever the tweet mentions multiple collectives besides the target of the hate message (like example \ref{example:collective3}). We think that the first case reveals an opportunity for improvement on the annotation process, where sometimes a collective might have been labeled using one word and other times using many.

Example \ref{example:property1} show an incorrect prediction on the Property done by a model trained following the experiment described in \ref{ref:single-predictions}. Although human trafficking could be considered as a negative consequence, the tweet does not explicitly associate it to a particular Collective. These models tend to identify phrases with negative connotations as Properties, disregarding if they are associated with the target group. This problem arises independently of the presence of a real Property and usually all words or phrases that could be considered as "negatives" are labeled by automatic predictors. Another error that automatic models are prone to are labeling bigger or smaller portions of text. Example \ref{example:property2} shows a prediction made by a model trained as described in section \ref{ref:mising-components}. The model correctly identified "illegally invade the U.S." as part of the Property, but missed the rest.

Regarding Pivots, we found that a common problem derivates from the incapacity of the models to jointly learn to find the pivot and the separation of the tweet into premises. Example \ref{example:pivot1} shows predictions made by a model trained following the settings described in \ref{ref:single-predictions} that found two words in different parts of the tweet that are directly related, but that are both within the justification, so they are not really a pivot between premises. A new setting for experimentation could provide the model with the information of where are the Justification and the Conclusion, and enforce to predict exactly one phrase within each of them. Another error found when predicting pivots comes from where multiple valid Pivots can be found within the premises. Example \ref{example:pivot2} shows prediction of a model also trained as described in \ref{ref:single-predictions} that found two valid Pivots: "Salvini-Salvini" and "invade-invasion". Each one of them could be considered a valid Pivot, though the only one that was labeled by the human annotator was "Salvini-Salvini". This phenomenon is related and could be considered as a consequence of the disagreement between annotators shown in the example \ref{fig:disagreement_pivot}. In order to avoid this kind of error, annotators should be instructed to label all the possible Pivots if there were more than one.

For Premises and Conclusions, we found also several cases where the model correctly divided the tweet in two premises but failed to assign the kind of the premise: if it was a Justification or a Conclusion. Example \ref{example:premise-conclusion} shows a prediction done by a model trained to jointly predict both Justification and Conclusion at the same time, as explained in \ref{ref:joint-predictions}. Here, the model correctly identifies both parts of the argument but fails to correctly assign the Justification and Conclusion in itself. It is interesting to note that models predicting a single component as described in \ref{ref:single-predictions} do the same mistake when predicting Justification and Conclusion for this same example. This correlates with similar discrepancies between annotators shown in example \ref{agreement:justification-conclusion}.

For the Types of premises, models trained following the settings described in \ref{ref:single-predictions} usually fail to predict the minority classes ('Value' for Conclusions and 'Value' and 'Policy' on Justifications). On the contrary, performance on these classes improves when models are trained following the settings described in \ref{ref:type-of-premise}. We found that using both kind of premises for training instead of just one no only increases the amount of examples but also leverages the distribution among classes, which leds to a significant boost in performance, as shown in table \ref{tab:results-premise-types}. Example \ref{example:type-of-premise-pred} shows a Justification predicted as Policy by a model trained using only justifications and then correctly predicted as Value by a model trained using both Justifications and Conclusions.

% \todo{add example of type of premise y comparar con ejemplo de experimento 5.4}

\begin{figure}[H]
    \texttt{\#Prayers for this mother \#NoIllegals \#SendThemAllBack w/ their families \#NoDACA \#BuildTheWallNow}
    \caption{Example of Non-Argumentative tweet incorrectly labeled as argumentative by automatic model. The tweet refers to a context that is missing on the text}
    \label{example:non-arg}

\end{figure}

\begin{figure}
    \texttt{Video: (part 1) London \#BNP a frame trailer with patriotic sound system on the road in and around our capital city "say no to \textbf{immigration}" \#Brexit \#Immigration \#ImmigrationBan \#London \#England \#BrexitBorder \#Brexiteer \#Brexiteers \#BrexitGoodNews \#BrexitChaos}
    \caption{Example of prediction of Collective from experiment described in \ref{ref:single-predictions}. Though the model finds a mention of a Collective that seems to be accurate, there is no explicit Property associated so it shoudn't have been labeled}
    \label{example:colective1}
\end{figure}

\begin{figure}
    \texttt{At this time, w-organized crime/returning \textbf{jihadists} it's a matter of national security. \#Italy \#Salvini must ignore international social engineers/cultural \textbf{marxists} \#V4 Itali  Kurz others must challenge empty threats from un-eu \textbf{\underline{migration}} pimps. What can they really do about it?}
    \caption{Model predicting only on tweets that have a Collective, besides correctly finding 'immigration', also labeled 'jihadists' and 'marxists', which are being used as properties for either the target collective or other groups (like 'international social engineers')}
    \label{example:collective3}
\end{figure}

\begin{figure}

    \texttt{\textbf{Chain \underline{migration}} imported 120K foreign nationals from terrorist-funding countries since 2005 - breitbart @user @user \#EndChainMigration \#EndDACA \#NoAmesty \#EndBirthrightCitizenshipForIllegalAliens \#BuildTheWall \#KeepAmericaSafe}
    \caption{Example of prediction of Collective from experiment described in \ref{ref:single-predictions}. The prediction seems to be accurate, but it included the word "chain" associated with migration. Differences like this arise whenever there are frequent phrases like "Chain Migration" or "Illegal immigrants"}
    \label{example:collective2}
\end{figure}

\begin{figure}[H]
    % \vspace{-0.1cm}
    % \centering
    % \begin{quote}
        \texttt{Please dont call it ""rescue"" - it's \textbf{human trafficking} \#PortsClosed \#SendThemBack \#BenefitSeekers}\\

        % \textbf{Predicted Property:}\hspace{0.7cm} "human trafficking"\\
        % \textbf{Real Property:}\hspace{1.6cm} None
    % \end{quote}
     % \vspace{-0.5cm}
    \caption{Example of prediction of Property. Predicted Property is bolded. There was no real property labeled in this example.}
    \label{example:property1}
    % \vspace{-0.6cm}

\end{figure}

\begin{figure}
    \texttt{@user the disgrace is the illegal parent who \underline{brought their kids on their cirme} \underline{spree to \textbf{illegally invade the U.S.}} \underline{so taxpayers pay for their kids} \underline{education wic and medicaid}. We don't owe illegals our tax dollars \#SendThemBack \#WalkAway \#Trump \#MAGA \#RedNationRising}
    \caption{Example of prediction of Property from experiment described in \ref{ref:mising-components}. Real Property is undelined while prediction is bolded. The model predicted just a portion of the real Property and left most of it unlabeled}
    \label{example:property2}
\end{figure}

\begin{figure}
        \texttt{\textbf{Americans agree with @user on immigration}. \underline{We can not afford to give welfare} \underline{to illegals while U.S. citizens} \underline{are homeless} \textbf{\#VoteDemsOut \#FamiliesBelongTogheterMarch}}
    \caption{Example of prediction of Conclusion. Real conclusion is underlined while predicted is bolded. Here, the two parts of the argument were correctly identified but predictor chose the conclusion incorrectly}
    \label{fig:my_label}
\end{figure}

\begin{figure}
        \texttt{\underline{Americans agree with} \underline{@user on immigration}. \textbf{We can not afford to give welfare to illegals while U.S. citizens are homeless} \underline{\#VoteDemsOut} \underline{\#FamiliesBelongTogheterMarch}}
    \caption{Example of prediction of Justification. Real justification is underlined, while predicted is bolded. Again, the two parts of the argument were correctly identified but predictor chose the incorrect half}
    \label{fig:my_label}
\end{figure}

\begin{figure}
    \texttt{\color{blue}Pressure on \textbf{Spain's} maritime border: Boatloads of \#Illegal \#Migrants Storm \textbf{Spanish} Tourist Beaches \& Scatter \color{red} \#StopTheInvasion \color{black} \#Unregistered \#UnVetted}
    \caption{Example of pivot predicted by model trained as described in section \ref{ref:single-predictions}. Justification is in blue, while conclusion is in red. Although the words selected establish a relation between themselves, they are both part of the justification, so they are not really a pivot between both premises}
    \label{example:pivot2}
\end{figure}

\begin{figure}
    \texttt{Rich African Countries don't take in African Migrants. Rich muslim countries don't take in muslim migrants. Rich latin american countries don't take it latin migrants. \textbf{But white countries are supposed to acept them??}}
    \caption{The conclusion (bolded) was predicted as Fact though it is a Policy}
    \label{example:type-of-conclusion1}
\end{figure}

\begin{figure}
    \texttt{Angry that UN @user does its job and checks Lebanon isn't coercing Syrian refugees into returning home, \textbf{Lebanon will stop giving residence permits to the agencys international staff}}
    \caption{This conclusion was predicted as Policy though it is a Fact}
    \label{example:type-of-conclusion2}
\end{figure}

\begin{figure}
    \texttt{@user Amen: \textbf{See 'Canada in Decay' by Ricardo Duchesne for the similar reality of Canada}. We are not nations of immigrants.}
    \caption{The justification (bolded) was predicted as Fact though it is a Policy}
    \label{example:type-of-justification1}
\end{figure}

\begin{figure}
    \texttt{Good news. We are against illegal immigrants}
    \caption{The justification (bolded) was predicted as Fact though it is a Value}
    \label{example:type-of-justification2}
\end{figure}

\begin{figure}
    \texttt{@user \color{blue}\underline{Immigration in a picture} \color{black}:link: \color{red}\textbf{Some basic truths: Access to White people is not a human right.}}
    \caption{Example of prediction of Justification and Conclusion. Predicted Conclusion is shown in blue while the real one is bolded. Predicted Justification is shown in red while the real one is underlined. Models were able to correctly divide the tweet in two premises but failed to correctly recognize Justification and Conclusion}
    \label{example:premise-conclusion}
\end{figure}

\begin{figure}
    \texttt{@user \underline{Immigration in a picture} \color{black}:link: \textbf{Some basic truths: Access to White people is not a human right.}}
    \caption{Example of prediction of Conclusion. Predicted Conclusion is underlined while the real one is bolded.}
    \label{example:premise-conclusion}
\end{figure}

\begin{figure}
    \texttt{@user \textbf{Immigration in a picture} \color{black}:link: \underline{Some basic truths: Access to} \underline{White people is not a human right.}}
    \caption{Example of prediction of Justification. Predicted Conclusion is underlined while the real one is bolded.}
    \label{example:premise-conclusion}
\end{figure}

\begin{figure}
    \texttt{I do not want those vile thugs in our country}
    \caption{Justification labeled as Value by human annotator. This premise was predicted as Policy by a model trained following the settings described in \ref{ref:single-predictions} and was correctly identified as Value by a model trained as described in \ref{ref:type-of-premise}}
    \label{example:type-of-premise-pred}
\end{figure}

\section{Argument annotated social media corpora}
\label{sec:corpora-social-media}

There exist several datasets with argument annotations, but only a few of them annotate arguments on Twitter.
DART relies on Argumentation Theory~\cite{rahwan2009argumentation} finding relationships between tweets as a single unit, considered to be arguments within an Abstract Argumentation Framework~\cite{Dung95}. Tweets are considered as argumentative if they express opinion or claims showing stance about a particular topic, and then they are defined according to how they interact with other tweet-arguments. The work of~\citet{dusmanu-etal-2017-argument} extends the \textit{\#Grexit} subset of DART (987 tweets) with another 900 labeled for argument detection and adds labels for factual arguments recognition and source identification. However, abstract frameworks do not consider the inner structure of arguments and
 are not useful in providing an argumentative analysis in the context of a single tweet.

\citet{schaefer-stede-2020-annotation} labeled 300 replies to context tweets about Climate Change in German language with claims and evidence to support the claims. This was later expanded to 1200 tweets and the annotation scheme was refined to focus on particular argument properties~\cite{schaefer_stede_2022}. This is the only work, to our knowledge, where spans are annotated within a tweet, but it is not a hate dataset and does not have domain specific information.% oriented to facilitate responses.

Finally, \citet{bhatti-etal-2021-argument} created a dataset of 24100 tweets 
%collected with the Twitter API, 
searching two hashtags supporting and attacking Planned Parenthood.
%The hashtags are assumed to be claims and the rest of the tweet is assigned one label indicating if they support the claim, either with or without reasons, or not. 
The whole tweet is assigned a single label (i.e., support or not the claim) and there is no argumentative structure segmentation within, so it is impossible to differentiate aspects of argumentative information.

%\todo{ALGO SOBRE LA IMPORTANCIA DE APLICAR ESQUEMAS ARGUMENTATIVOS A CASOS REALES DE USER GENERATED TEXT EN REDES SOCIALES Y LA DIFICULTAD QUE CONLLEVA}

% \todo{Add more debate about adequacy}

\end{document}